%% file: root.tex
\documentclass[letterpaper, 10 pt, conference]{ieeeconf}  

\IEEEoverridecommandlockouts                              

\makeatletter
\let\NAT@parse\undefined
\makeatother
\usepackage[numbers,sort&compress]{natbib}

\usepackage[pdftex]{graphicx}
\usepackage{amsmath}
\usepackage{amssymb}  
\usepackage{subfigure}
\usepackage{multirow}
\usepackage{array,booktabs}
\usepackage{diagbox}
\usepackage{balance}
\usepackage{xspace}
\usepackage{algorithm}
\usepackage{algpseudocode}
\usepackage{adjustbox}
\usepackage{romannum}

\usepackage[final]{hyperref}
\hypersetup{
 colorlinks=true,
 linkcolor=blue,
 filecolor=magenta,
 urlcolor=blue,
 citecolor=black
}
\usepackage{cleveref}
\crefname{figure}{Fig.}{Figs.}
\Crefname{figure}{Fig.}{Figs.}
\usepackage{rpm_SIunits}
\usepackage{rpm_acronyms}
\usepackage{rpm_math}
\usepackage{rpm_misc}
\usepackage{bbm}

\usepackage{textcomp}

\usepackage{soul,color}
\usepackage{lipsum}
\usepackage{dsfont}

\usepackage{xcolor}

\newcommand{\bl}[1]{{\textcolor{blue}{#1}}}

\usepackage{amssymb}
\usepackage{pifont}
\usepackage{tikz} 
\usepackage{caption}
\usepackage[font=scriptsize]{caption}
\title{\LARGE \bf HeLiOS: Heterogeneous LiDAR Place Recognition \\ via Overlap-based Learning and Local Spherical Transformer}     

\author{Minwoo Jung${}^{1}$, Sangwoo Jung${}^{1}$, Hyeonjae Gil${}^{1}$ and Ayoung Kim${}^{1*}$
\thanks{$^\dagger$This work was supported by the National Research Foundation of Korea (NRF) grant funded by the Korea government (MSIT)(No. RS-2024-00461409), and in part by Institute of Information and Communications Technology Planning and Evaluation
(IITP) grant funded by the Korea Government (MSIT)(No.2022-0-00480).}
\thanks{$^{1}$M. Jung, S. Jung, H. Gil and A. Kim are with the Dept. of Mechanical Engineering, SNU, Seoul, S. Korea {\tt\small [moonshot, dan0130, h.gil, ayoungk]@snu.ac.kr}}}

\begin{document}

\maketitle
\thispagestyle{empty}
\pagestyle{empty}

\input{0_abstract}
\input{1_introduction}
\input{2_relatedwork}
\input{3_method}
\input{4_experiment}
\input{5_conclusion}

\balance
\small
\bibliographystyle{IEEEtranN} 
\bibliography{string-short,references}

\end{document}

%% file: 0_abstract.tex
\begin{abstract}
LiDAR place recognition is a crucial module in localization that matches the current location with previously observed environments. Most existing approaches in LiDAR place recognition dominantly focus on the spinning type LiDAR to exploit its large FOV for matching. However, with the recent emergence of various LiDAR types, the importance of matching data across different LiDAR types has grown significantly—a challenge that has been largely overlooked for many years. To address these challenges, we introduce HeLiOS, a deep network tailored for heterogeneous LiDAR place recognition, which utilizes small local windows with spherical transformers and optimal transport-based cluster assignment for robust global descriptors. Our overlap-based data mining and guided-triplet loss overcome the limitations of traditional distance-based mining and discrete class constraints. HeLiOS is validated on public datasets, demonstrating performance in heterogeneous LiDAR place recognition while including an evaluation for long-term recognition, showcasing its ability to handle unseen LiDAR types. We release the HeLiOS code as an open source for the robotics community at \bl{https://github.com/minwoo0611/HeLiOS}.
\end{abstract}

%% file: 1_introduction.tex
\section{Introduction}
\label{sec:intro}

\acf{LPR} identifies whether a location was previously visited by comparing current scans to past ones from a pair of \ac{LiDAR} scans. 
Among many \ac{LiDAR} types, high-resolution spinning \ac{LiDAR}s have been the most popular choice \cite{kim2018scan, kim2021scan, xu2023ring++, luo2023bevplace} to handle occlusions and deliver extensive data from their 360-degree coverage and comprehensive information. 
However, the reliance on high-resolution systems has constrained the generalized solutions across various LiDAR types.



In this paper, we shift the focus from this popular 360$^\circ$ scanning \ac{LiDAR} to a diverse range of different \ac{LiDAR} types, highlighting challenges in heterogeneous \ac{LPR}.
For example, limited data from narrow \ac{FOV} solid-state \ac{LiDAR}s \cite{yuan2023std, yuan2024btc,kim2024narrowing} reintroduce complexity to the problem. 
Resolution difference among sensors raised an issue as sparsity caused the same structure to appear differently, making it challenging to use 2D convolution \cite{ma2023cvtnet, ma2022overlaptransformer} or methods that are adapted from \ac{VPR} \cite{shan2021robust}.
The new scanning patterns introduced by Livox and Robosense \cite{lin2020decentralized} capture the surroundings in a completely different manner. As a result, the data varies significantly from sensor to sensor, even when viewed from the same location.
These disparities pose significant challenges for heterogeneous \ac{LPR} \cite{jung2023helipr}.



Recently, transformer-based methods \cite{ma2022overlaptransformer, ma2023cvtnet} boosted performance to tackle similar challenges such as Ground-Aerial \ac{LiDAR} \cite{guan2023crossloc3d} or Camera-\ac{LiDAR} place recognition \cite{lee2023lc2, komorowski2021minkloc++}. However, applying transformers to heterogenous \ac{LiDAR}s often fails to encode data into a common embedding space due to different distributions, while using individual transformers for each source remains a limitation of generalizability. Handcrafted algorithms \cite{yuan2023std, yuan2024btc, kim2024narrowing} offer viable solutions, but they require multiple scans and specific poses. Furthermore, their applicability for heterogeneous \ac{LiDAR}s is still limited as the validation is only done within homogeneous \ac{LiDAR}s. 

\begin{figure}[!t]
    \centering
    \includegraphics[width=.95\columnwidth]{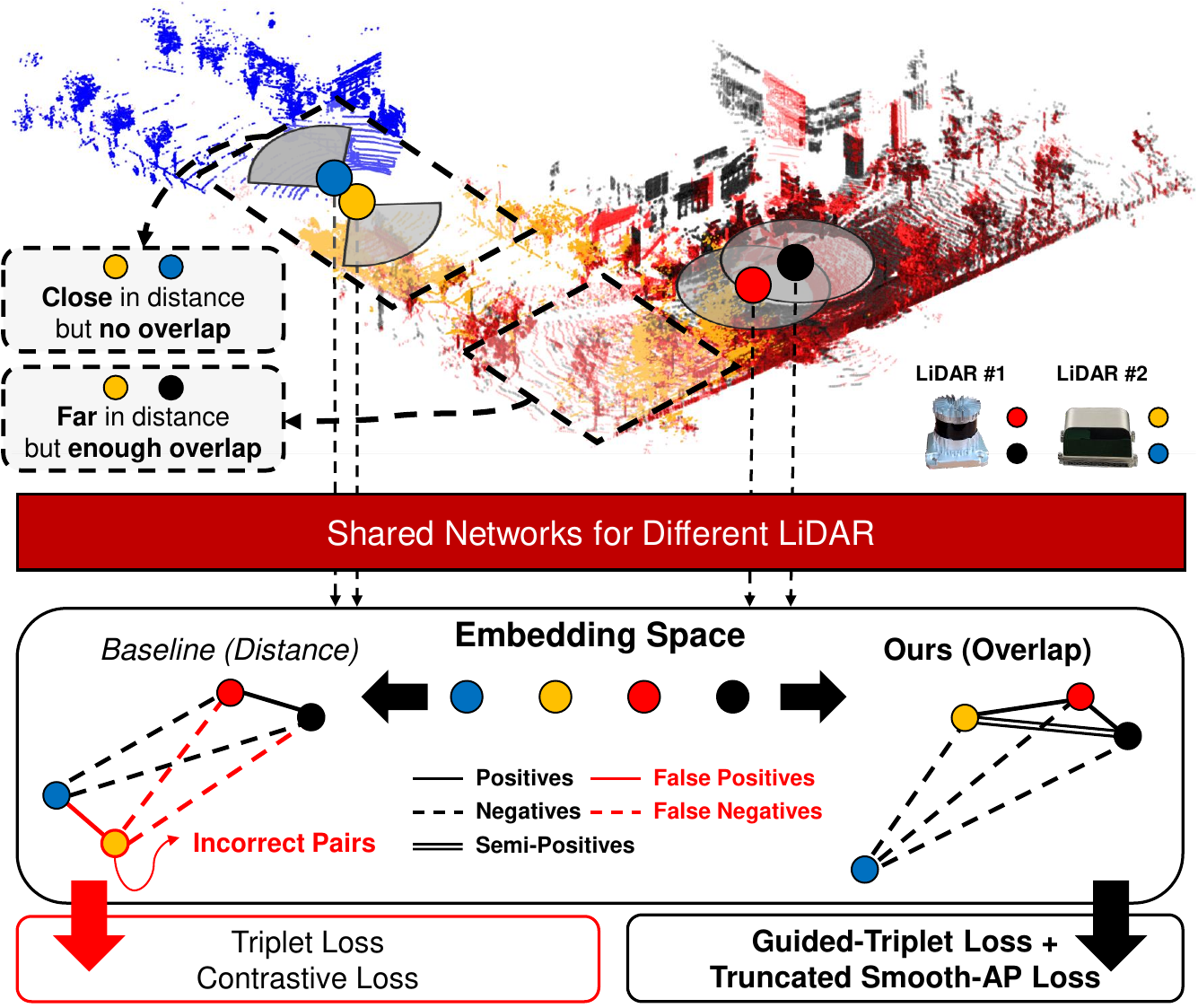}
    \caption{HeLiOS utilizes overlap for mining and the loss function. Conventional distance-based mining might lead to incorrect pairings, such as blue-yellow circles (close distance, no overlap) or black-yellow circles (far apart, overlap). Incorrect pairings can negatively impact both the training process and overall performance.}
    \label{fig:overall_diagram}
    \vspace{-8mm}
\end{figure}

In this paper, we present a novel network for single scan heterogeneous \ac{LPR}, that addresses variations in \ac{FOV}, resolution, and scanning patterns. We extend the spherical transformer \cite{lai2023spherical} along the radial direction to create smaller patches to better capture the point cloud's local distributions. Global descriptors are generated using optimal transport-based cluster assignment \cite{izquierdo2024optimal}, which filters out uninformative features while preserving the original distribution of local features. Moreover, our descriptor offers a flexible dimension range, allowing customization based on the user's objectives and requirements. An overlap-based data mining and \ac{TSAP} loss with guided-triplet loss are introduced to optimize the training process. The overlap-based approach addresses the inaccuracies associated with distance-based methods, providing more consistent constraints through a semi-positive class, as illustrated in \figref{fig:overall_diagram}. 

Our main contributions are as follows:

\begin{itemize}
    \item We introduce HeLiOS, a deep network to overcome the major challenges of heterogeneous \ac{LiDAR}s: diverse scanning patterns, \ac{FOV}s, and resolutions. Using the sparse convolution and spherical transformer with a local window, HeLiOS captures both low-level and high-level information. To our knowledge, this is the first method tailored for heterogeneous \ac{LiDAR} systems.
    \item We propose overlap-based data mining and guided-triplet loss to capture the relationships between \ac{LiDAR} descriptors, overcoming the limitations of discrete classes in traditional triplet loss and reducing wrong classes in distance-based mining. Our semi-positives ensure comprehensive constraints across various labels.
    \item HeLiOS is validated on public datasets, exhibiting superior performance in inter-\ac{LiDAR} and inter-session place recognition compared to \ac{SOTA} methods. We open-source HeLiOS for \ac{LPR}'s community. 
\end{itemize}

%% file: 2_relatedwork.tex
\section{related work}
\label{sec:relatedwork}
\subsection{Deep Learning in LiDAR Place Recognition}


\begin{figure*}[!t]
    \centering
    \includegraphics[width=0.89\textwidth]{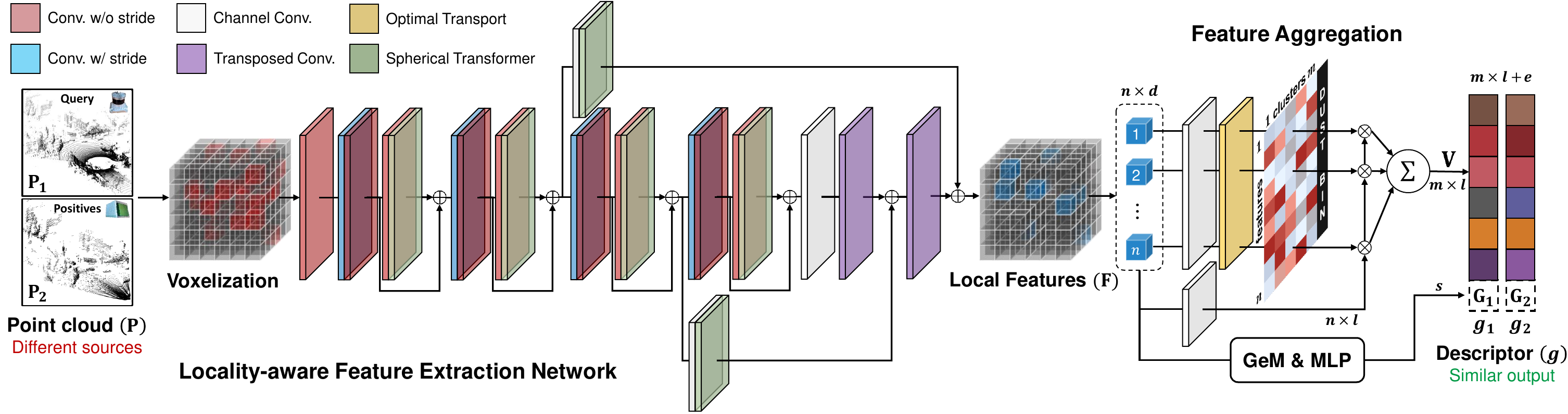}
    \caption{The overall pipeline of HeLiOS. HeLiOS voxelizes point clouds \(\mathbf{P}\) from heterogeneous LiDARs, which are then processed through a shared feature extraction network with sparse convolution and a spherical transformer. Local features \(\mathbf{F}\) are aggregated into global descriptors using GeM and SALAD to produce similar descriptors, which are subsequently utilized for training and evaluation.}
    \label{fig:overview}
    \vspace{-6mm}
    \label{fig:pipeline}
\end{figure*}

A seminal work in the learning-based \ac{LPR}, PointNetVLAD \cite{uy2018pointnetvlad} used PointNet \cite{qi2017pointnet} for feature extraction. Traditional methods \cite{liu2019lpd, guo2019local} relied on \ac{MLP}, enhancing global descriptors by improving local contextual relationships and minimizing information loss. To reduce computational costs and better encode point cloud, OverlapTransformer \cite{ma2022overlaptransformer} and CVTNet \cite{ma2023cvtnet} chose 2D convolutions on images of projected point cloud, but it struggles on various image formats of heterogeneous \ac{LiDAR}s. 
Other methods \cite{komorowski2022improving, vidanapathirana2022logg3d, xia2023casspr} applied sparse convolution for efficient 3D computation, but still requiring performance improvements.

Recent studies employed transformer \cite{ma2023cvtnet, xia2023casspr} to enhance \ac{LPR} performance. While these approaches show improvement for homogeneous \ac{LiDAR}s, they struggle with heterogeneous \ac{LPR} due to varying data distributions, complicating the effective learning of attention mechanisms. For example, SALSA \cite{goswami2024salsa} utilized SphereFormer \cite{lai2023spherical} for feature extraction with multi-head attention applied within spherical windows. However, the windows are too large to accurately capture the distinct distributions of different \ac{LiDAR}s, limiting their effectiveness in heterogeneous \ac{LPR}s.

Our model leverages sparse convolution and transformer but focuses on adaptation for heterogeneous \ac{LiDAR}s. We divide the spherical window into smaller regions based on spherical coordinates (\(r\), \(\theta\), \(\phi\)), allowing the transformer to learn local distributions. Additionally, our model uses a shared encoder to generate descriptors that are effective across different \ac{LiDAR}s, distinguishing it from methods focused on multi-modal place recognition \cite{lee2023lc2, zhou2023lcpr, komorowski2021minkloc++}. 
\subsection{Data Mining Strategies and Losses for Place Recognition}
Traditional \ac{LPR} relied on distance-based sampling to generate positive and negative samples for data mining. However, this approach faced challenges with narrow \ac{FOV} \ac{LiDAR}s, as scans from the same location may not overlap. \citeauthor{leyvavallina2023generalizedcontrastiveoptimizationsiamese}~\cite{leyvavallina2023generalizedcontrastiveoptimizationsiamese} used overlap for data mining, but their method, tailored for images and dense maps, is unsuitable for sparse \ac{LiDAR}s. Similarly, OverlapNet \cite{chen2020rss} requires the height and width of the range image, making it hard to decide the common format for different \ac{LiDAR}s. In contrast, our approach calculates overlap directly in 3D space considering scan density. We classify data into positive, semi-positive, and negative to position descriptors within embedding space to be more suitable for heterogeneous \ac{LPR}.

Moreover, \ac{LPR} traditionally used triplet loss \cite{dong2018triplet} and contrastive loss \cite{9577669}, which are designed for discrete class tasks like image classification \cite{Ge_2018_ECCV, Lee_2023_CVPR}, limiting their effectiveness in \ac{LPR}. \citeauthor{leyvavallina2023generalizedcontrastiveoptimizationsiamese}~\cite{leyvavallina2023generalizedcontrastiveoptimizationsiamese} improved this by multiplying overlap with contrastive loss, but sample with small overlap can still be embedded near the negatives. OverlapNet \cite{chen2020rss} utilized overlap in loss without affecting descriptor distribution. LoGG3D-Net \cite{vidanapathirana2022logg3d} introduced local consistency loss for better feature similarity but struggles with varying scanning patterns and density. MinkLoc3Dv2 \cite{komorowski2022improving} used \ac{TSAP} loss to improve average precision for top $k$ positives but lacks explicit distance constraints. We combine \ac{TSAP} loss with guided-triplet loss, adding overlap-based margin constraints to regulate descriptor distances better. 


%% file: 3_method.tex
\section{Methodology}
\subsection{Problem Definition}
\ac{LPR} aims to generate a global descriptor from a point cloud \(\mathbf{P} \in \mathbb{R}^{N \times 3}\), where the \(N\) points are defined by its spatial coordinates \((x, y, z)\). To encode the point cloud into a descriptor, a mapping function \(\Omega = h(f(\cdot))\) is employed, where the feature extraction function \(f(\cdot): \mathbb{R}^{N \times 3} \rightarrow \mathbb{R}^{n \times d}\) derives local embeddings from \(N\) points, and the aggregation function \(h(\cdot): \mathbb{R}^{n \times d} \rightarrow \mathbb{R}^e\) compresses these embeddings into a global descriptor \(g \in \mathbb{R}^{e}\) of fixed dimensions. The goal is to optimize \(\Omega\) to meet the following conditions:
\begin{eqnarray}
    \mathcal{D}(\mathbf{x}_q, \mathbf{x}_i) \ \mathcal{D}(\mathbf{x}_q, \mathbf{x}_j) \implies d_g(g_q, g_i) < d_g(g_q, g_j),
\end{eqnarray}
while \(\mathbf{x}\) denotes the locations of the point cloud, \(\mathcal{D}(\cdot)\) represents the distance between the locations, and \(d_g(\cdot)\) signifies the distance within the embedding space. Optimization of \(\Omega\) is achieved through metric learning, applying a loss function to the global descriptor derived from the training set.


\begin{figure}[!t]
    \centering
    \includegraphics[width=0.85\columnwidth]{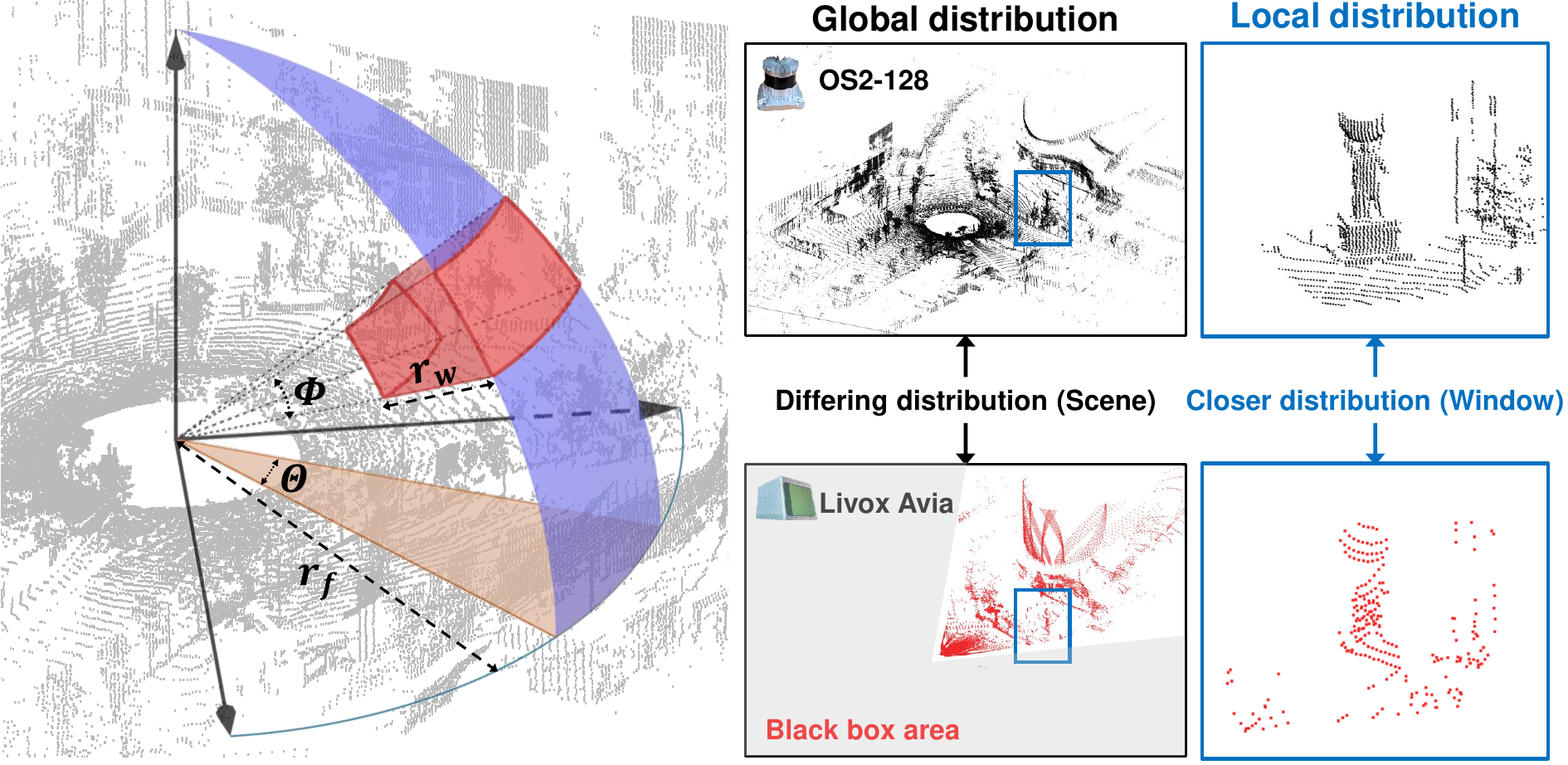}
    \caption{Local spherical window for applying multi-head attention in heterogeneous \ac{LPR}. Due to the differing global distribution when scanning the entire scene with different \ac{LiDAR}s, training attention is challenging. Distribution is closer within smaller, localized windows, enabling effective attention learning.}
    \vspace{-7mm}
    \label{fig:sphere}
\end{figure}

\subsection{Locality-aware Feature Extraction Network}
To encode local features from point clouds, we utilize a U-Net style architecture with sparse convolution \cite{choy20194d}. Point clouds are voxelized along the (\(x\), \(y\), \(z\)) axes, ensuring uniform resolution in 3D space. While sparse convolution effectively captures local information within each voxel, the variations in coverage and scanning patterns across heterogeneous \ac{LiDAR}s present significant challenges. The differing distribution of heterogeneous \ac{LiDAR}s disturbs their application by potentially causing divergence during training, even with transformers or self-attention that provide a viable solution for embedding point clouds into a common space. 

To focus on local distribution rather than global distribution, we divide the space with a spherical window and leverage multi-head attention as \figref{fig:sphere}. Inspired by vision transformer \cite{han2022survey} that segment images into patches, we apply multi-head attention to 3D voxels within spherical windows defined by spherical coordinates \((r, \theta, \phi)\). As the volume of spherical windows increases with $r$ for the same $\theta$ and $\phi$, multi-head attention using a consistent cubic window is also applied to complement local information. Consequently, each half of the multi-head attention output is derived from cubic and spherical windows. This approach differs from SphereFormer \cite{lai2023spherical}, which uses full-radius spherical windows.

We apply the spherical transformer only where skip connections exist, allowing attention to be processed while preserving the output from the sparse convolution layer. Additionally, as the voxels are progressively downsampled through the model, some windows may not contain enough voxels. To address this, when each time sparse convolution with stride is applied, the spherical window scale is expanded by 1.5, while the cubic window scale remains constant. Our feature extraction network pipeline is illustrated in \figref{fig:pipeline}.

\subsection{Feature Aggregation with optimal transport}

Different \ac{LiDAR}s generate varying numbers of local features \(\mathbf{F} \in \mathbb{R}^{n \times d}\), complicating feature aggregation. To address this, we adapt a clustering-based approach that is less sensitive to these variations. Specifically, we adapt SALAD \cite{izquierdo2024optimal} from vision tasks, where image patches are used as input. We consider voxels with local features as patches, enabling us to apply a method designed for a different task.

\(\mathbf{F}\) is processed through a channel-wise convolution layer to predict the score matrix, \(\mathbf{S} \in \mathbb{R}^{n \times m}\), where \(m\) is the cluster number. To manage non-informative points, a dustbin column is added, modifying the score matrix to \(\bar{\mathbf{S}} \in \mathbb{R}^{n \times (m + 1)}\). The Sinkhorn algorithm \cite{NIPS2013_af21d0c9} is then applied to optimize the feature-to-cluster assignment, creating a refined score matrix \(\mathbf{R} \in \mathbb{R}^{n \times m}\) obtained by iteratively normalizing the rows and columns of $\exp ({\bar{\mathbf{S}}})$ and dropping the dustbin. To align features with the cluster space, \(\mathbf{F}\) is projected to a lower-dimensional \(\bar{\mathbf{F}} \in \mathbb{R}^{n \times l}\). Finally, the aggregated feature matrix \(\mathbf{V} \in \mathbb{R}^{m \times l}\) is computed, where \(V_{j,k}\) is:
\begin{eqnarray}
V_{j,k} &=& \sum_{i=1}^{n} R_{i,k} \cdot \bar{F}_{i,j}
\end{eqnarray} 

To address the loss of global context that may occur during the clustering process, we also employ GeM pooling \cite{radenovic2018fine} combined with MLP layers. This produces a compact global representation \(\mathbf{G} \in \mathbb{R}^{e}\). The final descriptor is created by concatenating the flattened global features \(\mathbf{G}\) with the aggregated features \(\mathbf{V}\), ensuring a comprehensive representation with minimal dimensional increase. The descriptor \(g \in \mathbb{R}^{m \times l + e}\) is then utilized for both training and evaluation.

Unlike conventional place recognition that utilizes 256 or 512 descriptor dimension, HeLiOS exploits the various dimension based on \(m\), \(l\) and \(e\). This approach offers flexibility in adjusting the dimension, allowing users to balance computational requirements and performance by customizing the trade-off between time complexity and accuracy.

\subsection{Overlap Guided Metric Learning}

\subsubsection{Overlap-Based Data Mining}
\begin{figure}[!t]
    \subfigure[Overlap Calculation]{        
        \includegraphics[width=0.55\columnwidth]{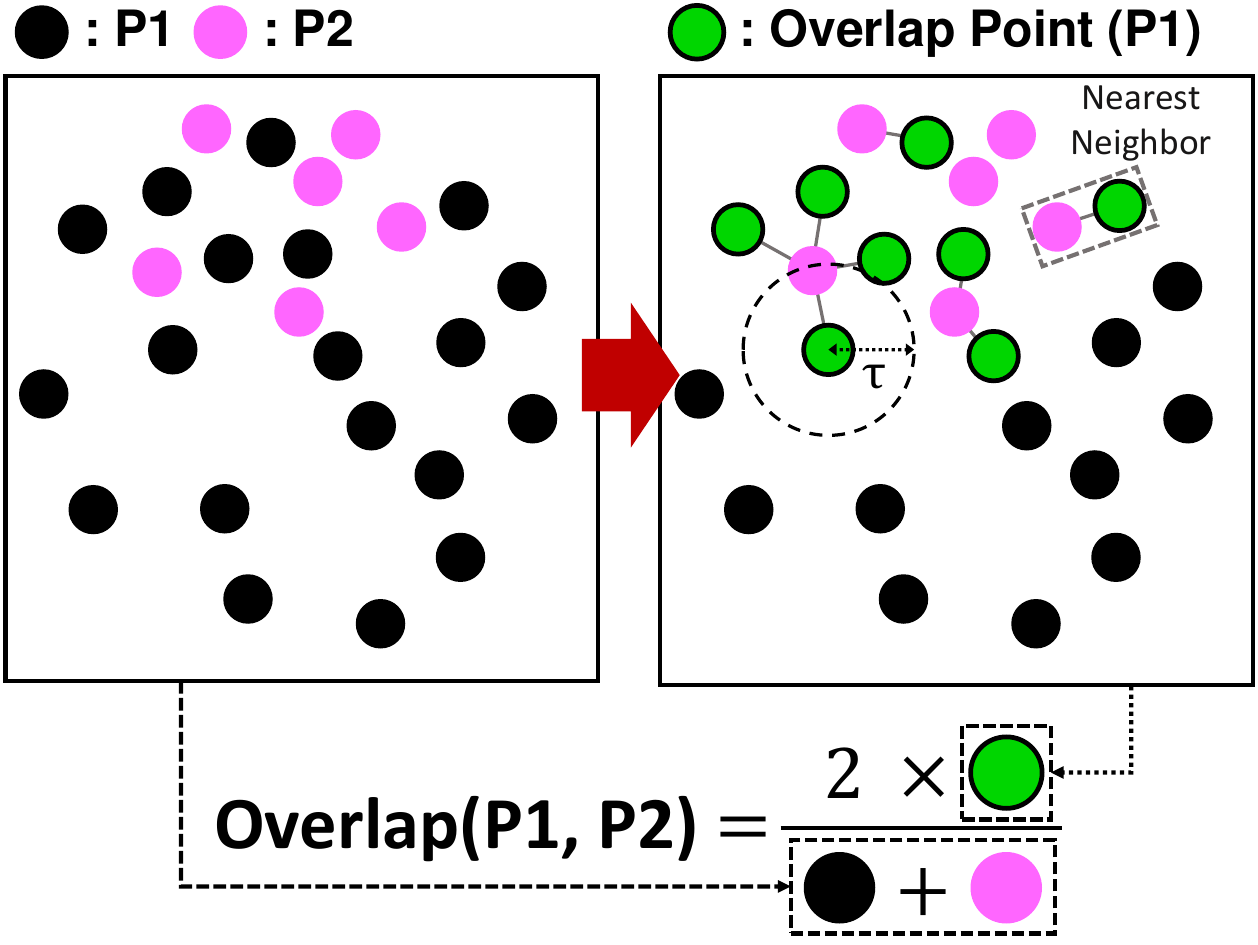}
        \label{fig:overlap}
        }
    \subfigure[Real-World Example]{        
        \includegraphics[width=0.3\columnwidth]{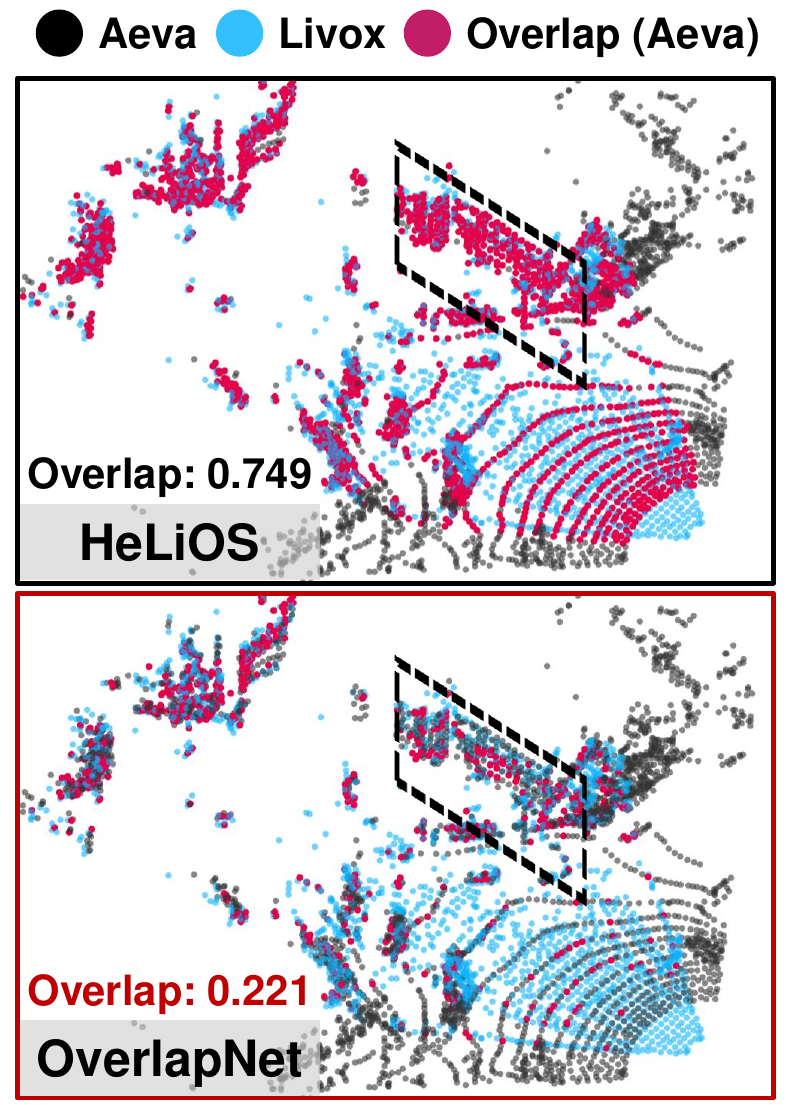}
        \label{fig:overlap_ex}
        }
    \vspace{-2mm}
    \caption{(a) Overlap calculation to illustrate \eqref{eq:overlap} as a diagram. (b) HeLiOS computes overlap for different LiDARs in 3D space. In contrast, OverlapNet misrepresents real-world overlap even if the \ac{LiDAR}s are in the same location, as their overlap occurs only when each point falls into the same pixel in a range image.} 
    \vspace{-7mm}
\end{figure}

Traditional metric learning for place recognition often relies on distance-based sampling, which can fail with heterogeneous \ac{LiDAR}s as \figref{fig:overall_diagram}. The naive distance-based sampling may result in unrelated positives or negatives, requiring a more refined approach considering \ac{LiDAR}-specific characteristics. To address this, we employ an overlap-based data mining method, where defining the overlap between point clouds, \(P_1\) and \(P_2\), as:
\begin{eqnarray}
\label{eq:overlap}
    \hat{\text{O}}(P_1, P_2) &=& \frac{2 \times \sum_{i=1}^{N_1} \mathds{1} \left(\text{NN}(P_1^i, P_2) < \tau\right)}{N_1 + N_2},
\end{eqnarray}
where \(\text{NN}\) returns the distance to the nearest neighbor in the other point cloud, and \(\mathds{1}(\cdot)\) is an indicator function. The overall process can be found in \figref{fig:overlap}. To reduce the computationally expensive \(n \times n\) overlap matrix calculation, where \(n\) is the number of samples, we truncate the overlap calculation if the distance exceeds twice the maximum scan range. Furthermore, point clouds are voxelized with size $\delta$ to reduce computational costs and standardize resolution, preventing the large overlap in dense regions. We set \(\tau = 1.5\delta\) to ensure robustness to minor misalignment.

The use of the maximum overlap value ensures stable and consistent learning, keeping the overlap constant between the point clouds in a reversed relation. Final overlap is computed as $\text{O}(P_1, P_2) = \max(\hat{\text{O}}(P_1, P_2), \hat{\text{O}}(P_2, P_1))$ with a maximum threshold of 1. Our method addresses challenges in determining overlap with range images of different \ac{LiDAR}s, such as varying resolutions \cite{chen2020rss} and the sparsity of \ac{LiDAR} in 3D space \cite{leyvavallina2023generalizedcontrastiveoptimizationsiamese}, providing a reliable overlap measurement as \figref{fig:overlap_ex}. Scans are categorized as positive if overlap is over 0.5, semi-positive between 0 and 0.5, and negative if zero, enhancing robustness and generalization in metric learning.

\subsubsection{Guided-Triplet Loss}
In \ac{LPR}, the balance of ranking relevant scans and controlling their distances in the embedding space is crucial for robust model. The \ac{TSAP} loss \(\mathcal{L_{TSAP}}\) \cite{komorowski2022improving} ranks the top \(k\) positives but lacks explicit control over distances between positive (${p}$) and negative (${n}$) from the query (${q}$). This leads to a dispersed distribution of descriptors with poorly defined boundaries, reducing the discriminative power and causing instability during training.

To address these limitations, we introduce a combined loss function that incorporates both \(\mathcal{L_{TSAP}}\) and a guided-triplet loss \(\mathcal{L_{GT}}\). Unlike triplet loss \(\mathcal{L_{T}}\) with a fixed margin to separate two classes, guided-triplet loss employs adaptive margins based on the overlap to allow more general regulation of distances. This reflects varying degrees of similarity between scans and ensures that the embeddings are more distributed and distances are effectively controlled. Additionally, we incorporate two \(\mathcal{L_{GT}}\) based on the relationship with semi-positives (${s}$) and the others. Semi-positive should be closer to positives but still separated from them and positioned far from negatives. Guided-triplet loss \(\mathcal{L_{GT}}\) is formulated as:
\begin{eqnarray}
\mathcal{L_{GT}}(q, u, v) = \max( d_g(q, u) - d_g(q, v) + \alpha_{uv}, 0), 
\end{eqnarray}
where \(d_g(\cdot)\) are distances in the embedding space, and \(\alpha_{uv}\) is the adaptive margin based on overlap. The adaptive margin \(\alpha_{ps}\) for positive and semi-positive pair, and \(\alpha_{sn}\) for semi-positive and negative pair are defined as:
\begin{flalign}
\label{eq:adapt}
& \alpha_{uv} = 
\begin{cases} 
m_1 \cdot (\text{OV}(q,p) - \text{OV}(q,s)) \hspace{0.3mm} ~~~~~~~\text{if } (u = p, v = s) \\
m_2 \cdot (\text{OV}(q,s) - \text{OV}(q,n) + 1) ~~ \text{if } (u = s, v = n)
\end{cases} \nonumber \\
& \hspace{12.7mm} \text{OV}(q,u) \triangleq \log(\beta \cdot \text{O}(q,u) + 1), 
\end{flalign}
where \(m_1\), \(m_2\), and \(\beta\) are the scale factors, and logarithm regularizes the effect of overlap, ensuring large overlaps yield similar values while amplifying differences of small overlaps. 
In \eqref{eq:adapt}, we add \(\text{OV}(q,n)\) for the readability, even if it is always zero. To divide the semi-positive and negative more distinctly, an additional distance of \(m_2\) is provided for \(\alpha_{sn}\). 

By combining \(\mathcal{L_{TSAP}}\) with \(\mathcal{L_{GT}}\), our total loss function not only focuses on ranking the most relevant scans but also maintains appropriate distances between positives, semi-positives, and negatives. The total loss function is given as:
\begin{eqnarray}
\mathcal{L} = \mathcal{L_{TSAP}} + \omega_1 \cdot \mathcal{L_{GT}}(q, p, s) + \omega_2 \cdot \mathcal{L_{GT}}(q, s, n) ,
\end{eqnarray}
which enhances model generalization by ensuring well organization of global descriptors in the embedding space.

%% file: 4_experiment.tex
\input{tab/overall_results}
\section{experiment}
\label{sec:experiment}

\subsection{Implementation Details}
We trained HeLiOS on a GeForce RTX 3090 for 80 epochs using a MultiStepLR scheduler with an initial learning rate of 0.001. The maximum range is limited to 100\unit{m}, and the 8192 points from a single scan are normalized within \([-1, 1]\). The spherical windows are set to \((10\text{m}, 1.8^\circ, 1.8^\circ)\), and the voxel size \(d\) for overlap calculation is 4\(\meter\). For the guided-triplet loss, the weights \(\omega_1\) and \(\omega_2\) are set to 0.1, and the scaling factors \((m_1, m_2, \beta)\) are configured as \((0.02, 0.19, e-1)\).

\subsection{Datasets and Evaluation Metric}  
We evaluated HeLiOS on three public datasets: NCLT (HDL-32E) \cite{carlevaris2016university}, MulRan (OS1-64) \cite{kim2020mulran}, and HeLiPR (OS2-128, Livox Avia, Aeva Aeries II, and VLP-16C) \cite{jung2023helipr}. We compared HeLiOS against several methods, including PointNetVLAD \cite{uy2018pointnetvlad}, LoGG3D-Net \cite{vidanapathirana2022logg3d}, CASSPR \cite{xia2023casspr}, CROSSLOC3D \cite{guan2023crossloc3d}, MinkLoc3Dv2 \cite{komorowski2022improving}, and handcrafted descriptor SOLID \cite{kim2024narrowing}, applying our overlap criteria across all benchmarks for a fair comparison.

We chose Average Recall@k (AR@k) for evaluation. A retrieval is correct if the overlap between the query and retrieval exceeds 0.5. HeLiOS with parameters \((m, l, e) = (64, 128, 256)\) is evaluated to demonstrate its complex descriptor capability. For fairness, a smaller architecture, HeLiOS-S, is also configured with lightweight parameters \((m, l, e) = (8, 32, 0)\) for a descriptor dimension of 256.

\subsection{Heterogeneous LiDAR Place Recognition}
\label{sec:hetero}
The HeLiPR dataset is used for heterogeneous \ac{LPR}. Training is conducted on \texttt{DCC04-06}, \texttt{KAIST04-06}, and \texttt{Riverside04-06} with four \ac{LiDAR} types, sampled at $5\meter$ intervals, totaling $41k$ samples. Testing is conducted on  \texttt{Roundabout01-03}, \texttt{Town01-03}, \texttt{Bridge01} (paired with \texttt{04}), and \texttt{Bridge02} (paired with \texttt{03}) to ensure sufficient overlap, with $96k$ samples also sampled at $5\meter$.

Queries are classified by grouping Aeva and Livox as ``Narrow'' and Ouster and Velodyne as ``Wide''. The first sequences of Aeva and Ouster are chosen as databases, while all sequences and \ac{LiDAR}s at each location are queried against them. For example, to evaluate Ouster and ``Wide'' in \texttt{Roundabout}, the database is \texttt{Roundabout01-Ouster}, while results are averaged over the recalls of six cases (3 sequences $\times$ 2 \ac{LiDAR}s) to cover both intra-session and inter-session. Scans within 30 seconds of the query are excluded to prevent obvious matching occurs in intra-session. 

\begin{figure}[!t]
    \centering
    \includegraphics[width=0.98\columnwidth]{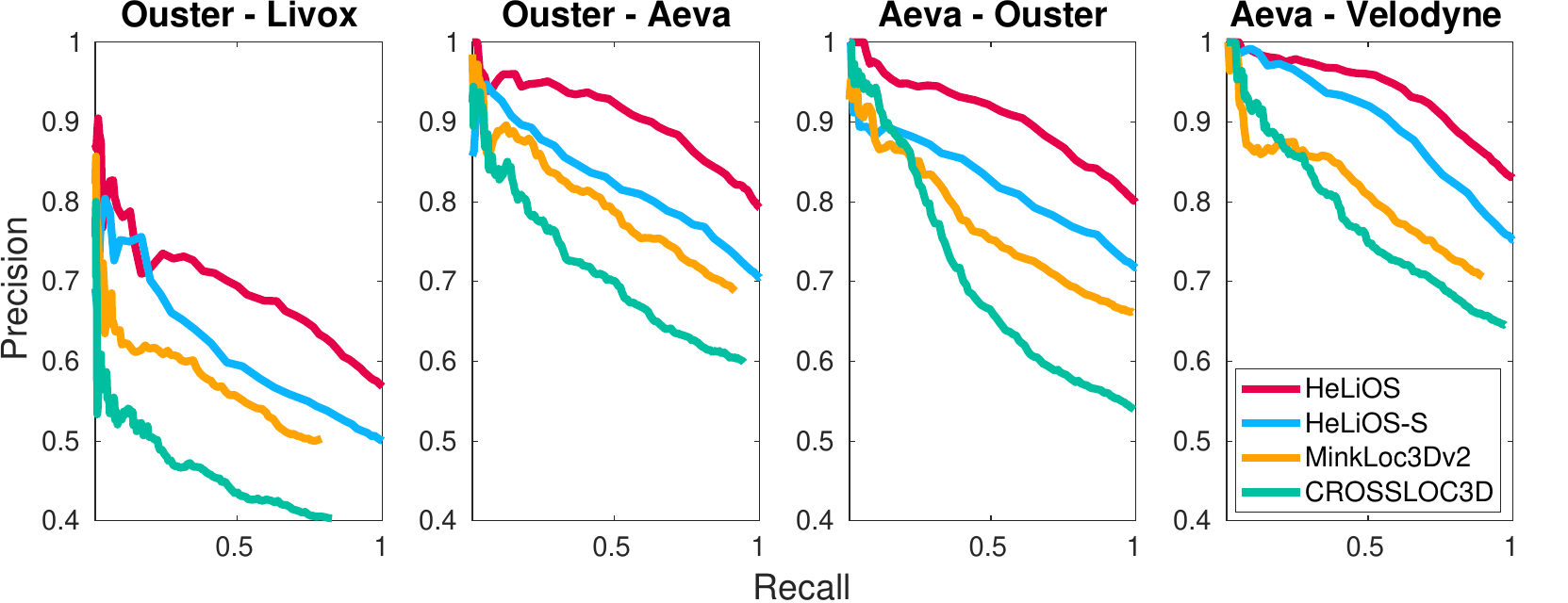}
    \caption{PR curves with heterogeneous \ac{LiDAR}s. The title of each curve represents the database from \texttt{Roundabout01} and the query from \texttt{Roundabout02}. HeLiOS surpasses other methods regardless of the size of the descriptor. 
    }
    \label{fig:PR-curve}
    \vspace{-3mm}
\end{figure}

\begin{figure}[!t]
    \centering
    \includegraphics[width=0.98\columnwidth]{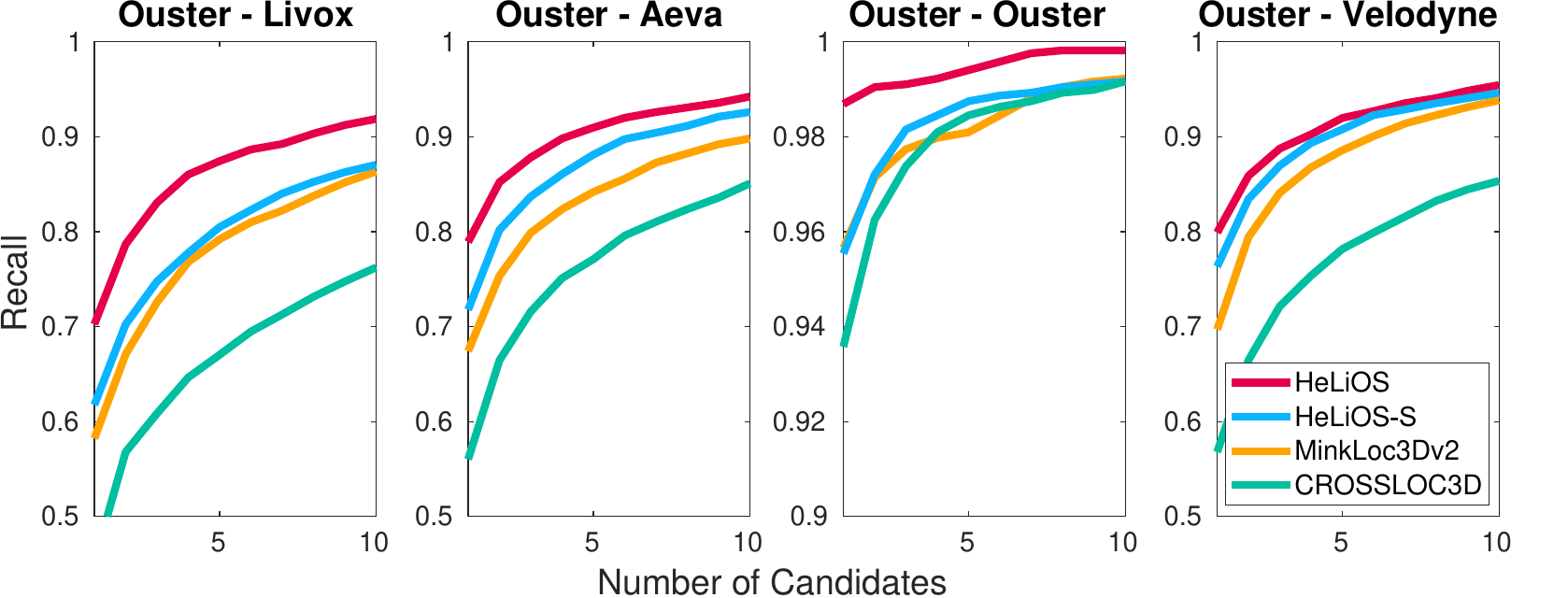}    
    \caption{R@N curves on \texttt{Town03} with heterogeneous LiDARs are shown, where each curve title indicates the database from \texttt{Town01} and the query from \texttt{Town03}. HeLiOS consistently outperforms other \ac{SOTA} methods when retrieving the top 10 neighbors.}
    \label{fig:recall}
    \vspace{-6mm}
\end{figure}

As seen in \tabref{tab:overall}, even with lightweight 256 dimensions, HeLiOS-S surpasses others in most cases. Thanks to the spherical transformer and overlap-based metric learning, HeLiOS significantly outperforms all methods across all places. Interestingly, MinkLoc3Dv2 achieves the third-best results despite using only convolutional layers. In contrast, CASSPR and CROSSLOC3D demonstrate weaker performance as transformers applied to entire point clouds struggle with the varying distributions of heterogeneous \ac{LiDAR}s. LoGG3D-Net delivers inferior results, as its local consistency loss fails to accommodate differing \ac{LiDAR} distributions.
SOLID exhibits limitations in its descriptors across different \ac{FOV}s.

We present the PR and AR@N curves for the top four methods in \figref{fig:PR-curve} and \figref{fig:recall}. The PR curves display results for Ouster with ``Narrow'' and Aeva with ``Wide''. HeLiOS outperforms others, achieving high recall and precision. The top 10 neighbors and their recall are plotted in \figref{fig:recall}, where HeLiOS excels in retrieving top candidates, with HeLiOS-S yielding better results even for Ouster. HeLiOS on homogeneous \ac{LiDAR} is further discussed in \secref{sec:homogeneous}.

\begin{figure}[!t]
    \centering
    \includegraphics[width=0.93\columnwidth]{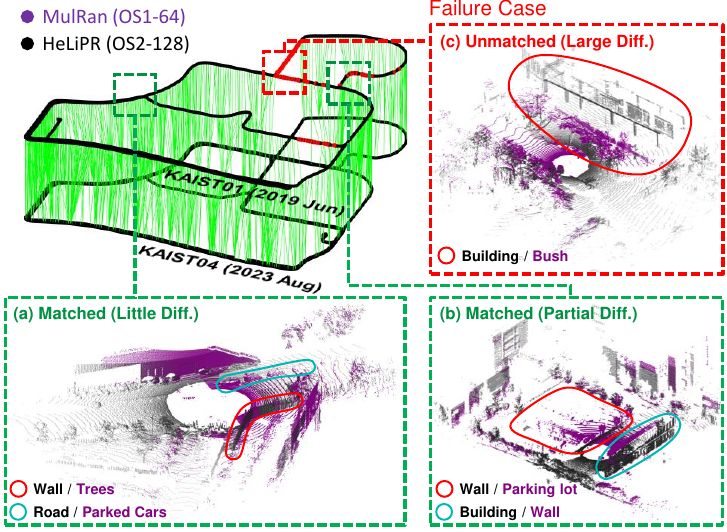}    
    \caption{Four-year long-term heterogeneous \ac{LPR} between MulRan (purple) and HeLiPR (black). Red and cyan indicate significant differences between point clouds. (a, b) HeLiOS matches most queries despite long-term changes, demonstrating robustness to scene variations. (c) A building (black) replacing bushes (purple) leads to failure due to the drastic change in the scene's appearance.
    }
    \label{fig:longterm}
    \vspace{-7mm}
\end{figure}

\subsection{Long-term Place Recognition with Heterogeneous LiDAR}
Long-term place recognition between HeLiPR and MulRan is evaluated. We set \texttt{KAIST01} from MulRan as the query and \texttt{KAIST04-Ouster} from HeLiPR as the database. Compared to \texttt{KAIST04}, a sequence from training, \texttt{KAIST01} is an unseen dataset captured with a different system and \ac{LiDAR} (OS1-64) with different \ac{FOV} and occlusion. For precise verification, the two sequences are aligned using ELite \cite{hjgil-2025-icra} based on their respective ground truths. Despite a four-year gap and scene changes, HeLiOS successfully matches almost every query, as shown in \figref{fig:longterm}. Although some areas fail to retrieve, this occurs from entire scene differences between the query and database, resulting in distinct descriptors that understandably do not match. This demonstrates HeLiOS's capability for long-term place recognition and retrieving scans from unseen datasets and \ac{LiDAR}s in partial differences.

\subsection{Homogeneous LiDAR Place Recognition}
\label{sec:homogeneous}
Differing from \secref{sec:hetero}, the networks are trained and evaluated with homogeneous LiDAR data.
For the NCLT dataset, networks are trained on 8 sequences totaling $9.0k$ samples and tested with data from \texttt{2012-11-16} as the database and \texttt{2012-12-01} as the query with $1.8k$ samples. For the HeLiPR dataset, training is done separately on Livox and Aeva data, each totaling $10.3k$ samples. Place recognition is performed with \texttt{Roundabout01} as the database (\texttt{R01}) and \texttt{Roundabout03} as the query (\texttt{R03}), resulting in $3.4k$ test samples for each \ac{LiDAR}. Retrieval is considered correct if the overlap exceeds 0.8, as homogeneous \ac{LiDAR}s typically yield higher overlap than heterogeneous ones. As \tabref{tab:single}, HeLiOS-S achieves comparable performance with others despite not being designed for homogeneous \ac{LiDAR}. 

\input{tab/ablation_single}
\input{tab/ablation_loss}

\subsection{Ablation Studies}
We conducted ablation studies to highlight the performance differences and verify the impact of our proposed modules on HeLiOS. Training setup follows \secref{sec:hetero}, while results are average recall of both ``Narrow'' and ``Wide'' cases for \texttt{Roundabout} and \texttt{Town}.

\textbf{Effect of Loss Function:}
We assessed the effect of \(\mathcal{L_{GT}}\) by keeping \(\mathcal{L_{TSAP}}\) fixed for positive and negative pairs while varying the loss for \((s, n)\) and \((p, s)\) pairs, with \(\omega_1\) and \(\omega_2\) set to 0.1. As \tabref{tab:loss}, applying \(\mathcal{L_{TSAP}}\) to both class pairs only pushes the embeddings further apart, similar to pushing query-negative pairs away from query-positive pairs while leading to inaccuracy than applying only to \((p, n)\) pairs. Conversely, using \(\mathcal{L_{T}}\) for \((s, n)\) and \((p, s)\) allows for additional performance improvements by providing distance control not achievable with \(\mathcal{L_{TSAP}}\) alone. Proposed \(\mathcal{L_{GT}}\) with overlap-based adaptive margins outperforms \(\mathcal{L_{T}}\) with fixed margins. This shows that the adaptive margin of $\mathcal{L_{GT}}$ enhances the traditional discrete class separation, allowing the embedding process to reflect the real-world better.

\textbf{Variation in Spherical Transformer:}
We evaluated the effect of spherical transformer under different configurations: without the transformer, varying radial window size, and applying expansion. \tabref{tab:sphere} shows result with \(r_f\) and \(r_w\) of 100$\meter$ and 10$\meter$, and an expansion factor of 1.5. Larger windows (\(r_f\)) result in slight performance improvements over the no transformer case, suggesting larger windows dilute attention on local patterns due to varying distributions of heterogeneous \ac{LiDAR}s. Conversely, smaller windows (\(r_w\)) lead to significant performance gains as the network better captures local features while reducing the impact of differing distribution. Gradual expansion ensures local windows contain enough points for attention even in deeper layers, enhancing spatial information encoding of the network.
\input{tab/ablation_network}

\input{tab/ablation_dim}

\textbf{Multi Scalability:}
As \tabref{tab:dim}, the effect of descriptor dimensions on performance, time complexity, and using GeM in feature aggregation are examined. Runtime is split into inference time (including preprocessing) and retrieval time based on a database of approximately $1.6k$ samples. A larger dimension improves recall while inference time remains almost constant as descriptor expansion is managed within the aggregator; however, retrieval time increases due to distance calculations. All configures achieve real-time speed below the 10Hz \ac{LiDAR} frequency. \(e = 256\) configuration, including GeM, yields notable gains with minimal dimension expansion. Notably, (32, 64, 256) performs similarly to (64, 128, 0), and even higher dimensions like (64, 128, 0) is benefited from additional concatenation. This indicates that combining GeM and MLP effectively compensates for information loss during clustering assignments as the class token in the vision transformer, enhancing the global descriptor with the addition of smaller dimensions.

%% file: tab/overall_results.tex
\begin{table*}[]
\caption{Performance Comparison with Heterogeneous \ac{LiDAR}s (\textbf{\color[HTML]{FF0000}Red}: Best, {\color[HTML]{0000FF}Blue}: Second Best)}
\label{tab:overall}
\resizebox{0.98\textwidth}{!}{%
\begin{tabular}{lccccccccccccccccc} \toprule
 &
   &
  \multicolumn{4}{c}{\texttt{Roundabout}} &
  \multicolumn{4}{c}{\texttt{Town}} &
  \multicolumn{4}{c}{\texttt{Bridge01}} &
  \multicolumn{4}{c}{\texttt{Bridge02}} \\
 &
   &
  \multicolumn{2}{c}{Narrow} &
  \multicolumn{2}{c}{Wide} &
  \multicolumn{2}{c}{Narrow} &
  \multicolumn{2}{c}{Wide} &
  \multicolumn{2}{c}{Narrow} &
  \multicolumn{2}{c}{Wide} &
  \multicolumn{2}{c}{Narrow} &
  \multicolumn{2}{c}{Wide} \\ \cline{3-18} \rule{0pt}{2.5ex}
\multirow{-3}{*}{DB} &
  \multirow{-3}{*}{Methods} & 
  \multicolumn{1}{l}{AR@1} &
  \multicolumn{1}{l}{AR@5} &
  \multicolumn{1}{l}{AR@1} &
  \multicolumn{1}{l}{AR@5} &
  \multicolumn{1}{l}{AR@1} &
  \multicolumn{1}{l}{AR@5} &
  \multicolumn{1}{l}{AR@1} &
  \multicolumn{1}{l}{AR@5} &
  \multicolumn{1}{l}{AR@1} &
  \multicolumn{1}{l}{AR@5} &
  \multicolumn{1}{l}{AR@1} &
  \multicolumn{1}{l}{AR@5} &
  \multicolumn{1}{l}{AR@1} &
  \multicolumn{1}{l}{AR@5} &
  \multicolumn{1}{l}{AR@1} &
  \multicolumn{1}{l}{AR@5} \\ \midrule
\multicolumn{1}{l|}{} &
  \multicolumn{1}{l|}{SOLID \cite{kim2024narrowing}} &
  0.054 &
  0.146 &
  0.278 &
  0.386 &
  0.124 &
  0.300 &
  0.292 &
  0.404 &
  0.027 &
  0.073 &
  0.321 &
  0.394 &
  0.038 &
  0.098 &
  0.293 &
  0.371 \\
\multicolumn{1}{l|}{} &
  \multicolumn{1}{l|}{PointNetVLAD \cite{uy2018pointnetvlad}} &
  0.297 &
  0.518 &
  0.586 &
  0.744 &
  0.291 &
  0.530 &
  0.474 &
  0.664 &
  0.329 &
  0.495 &
  0.704 &
  0.823 &
  0.362 &
  0.558 &
  0.646 &
  0.754 \\
\multicolumn{1}{l|}{} &
  \multicolumn{1}{l|}{LoGG3D-Net \cite{vidanapathirana2022logg3d}} &
  0.012 &
  0.038 &
  0.151 &
  0.295 &
  0.011 &
  0.049 &
  0.098 &
  0.241 &
  0.027 &
  0.032 &
  0.219 &
  0.320 &
  0.012 &
  0.047 &
  0.186 &
  0.373 \\
\multicolumn{1}{l|}{} &
  \multicolumn{1}{l|}{{\color[HTML]{000000} CASSPR \cite{xia2023casspr}}} &
  {\color[HTML]{000000} 0.182} &
  {\color[HTML]{000000} 0.407} &
  {\color[HTML]{000000} 0.478} &
  0.703 &
  0.178 &
  0.418 &
  0.376 &
  0.598 &
  0.220 &
  0.423 &
  0.548 &
  0.702 &
  0.234 &
  0.482 &
  0.488 &
  0.634 \\
\multicolumn{1}{l|}{} &
  \multicolumn{1}{l|}{CROSSLOC3D \cite{guan2023crossloc3d}} &
  {\color[HTML]{000000} 0.499} &
  {\color[HTML]{000000} 0.785} &
  0.796 &
  0.876 &
  0.565 &
  0.785 &
  0.739 &
  0.876 &
  0.547 &
  0.745 &
  0.810 &
  0.906 &
  0.555 &
  0.756 &
  0.781 &
  0.891 \\ 
\multicolumn{1}{l|}{} & 
  \multicolumn{1}{l|}{{\color[HTML]{000000} MinkLoc3Dv2 \cite{komorowski2022improving}}} &
  {{0.620}} &
  {{0.790}} &
  {{0.870}} &
  {0.928} &
  {0.660} &
  {0.838} &
  {0.817} &
  {\color[HTML]{0000FF} 0.919} &
  {0.650} &
  {0.818} &
  {0.876} &
  {0.938} &
  {0.631} &
  {0.819} &
  {\color[HTML]{0000FF} 0.833} &
  {0.910} \\
\multicolumn{1}{l|}{} &
  \multicolumn{1}{l|}{HeLiOS-S} &
  {\color[HTML]{0000FF} 0.637} &
  {\color[HTML]{0000FF} 0.801} &
  {\color[HTML]{0000FF} 0.880} &
  {\color[HTML]{0000FF} 0.929} &
  {\color[HTML]{0000FF} 0.686} &
  {\color[HTML]{0000FF} 0.862} &
  {\color[HTML]{0000FF} 0.828} &
  {0.918} &
  {\color[HTML]{0000FF} 0.660} &
  {\color[HTML]{0000FF} 0.828} &
  {\color[HTML]{0000FF} 0.880} &
  {\color[HTML]{0000FF} 0.950} &
  {\color[HTML]{0000FF} 0.649} &
  {\color[HTML]{0000FF} 0.827} &
  {0.831} &
  {\color[HTML]{0000FF} 0.921} \\ 

\multicolumn{1}{c|}{\multirow{-8}{*}{\rotatebox{90}{Ouster}}} &
  \multicolumn{1}{l|}{HeLiOS} &
  {\color[HTML]{FF0000} \textbf{0.700}} &
  {\color[HTML]{FF0000} \textbf{0.852}} &
  {\color[HTML]{FF0000} \textbf{0.912}} &
  {\color[HTML]{FF0000} \textbf{0.946}} &
  {\color[HTML]{FF0000} \textbf{0.753}} &
  {\color[HTML]{FF0000} \textbf{0.903}} &
  {\color[HTML]{FF0000} \textbf{0.871}} &
  {\color[HTML]{FF0000} \textbf{0.937}} &
  {\color[HTML]{FF0000} \textbf{0.693}} &
  {\color[HTML]{FF0000} \textbf{0.853}} &
  {\color[HTML]{FF0000} \textbf{0.912}} &
  {\color[HTML]{FF0000} \textbf{0.969}} &
  {\color[HTML]{FF0000} \textbf{0.681}} &
  {\color[HTML]{FF0000} \textbf{0.849}} &
  {\color[HTML]{FF0000} \textbf{0.874}} &
  {\color[HTML]{FF0000} \textbf{0.950}} \\ \midrule
\multicolumn{1}{l|}{} &  
  \multicolumn{1}{l|}{SOLID \cite {kim2024narrowing}} &
  0.241 &
  0.442 &
  0.018 &
  0.129 &
  0.200 &
  0.346 &
  0.048 &
  0.195 &
  0.236 &
  0.332 &
  0.013 &
  0.048 &
  0.234 &
  0.340 &
  0.017 &
  0.058 \\
\multicolumn{1}{l|}{} &
  \multicolumn{1}{l|}{PointNetVLAD \cite{uy2018pointnetvlad}} &
  0.477 &
  0.624 &
  0.338 &
  0.591 &
  0.408 &
  0.588 &
  0.313 &
  0.573 &
  0.678 &
  0.821 &
  0.366 &
  0.598 &
  0.625 &
  0.777 &
  0.365 &
  0.609 \\
\multicolumn{1}{l|}{} &
  \multicolumn{1}{l|}{LoGG3D-Net \cite{vidanapathirana2022logg3d}} &
  0.022 &
  0.094 &
  0.029 &
  0.096 &
  0.033 &
  0.116 &
  0.026 &
  0.101 &
  0.015 &
  0.062 &
  0.011 &
  0.052 &
  0.026 &
  0.098 &
  0.025 &
  0.249 \\
\multicolumn{1}{l|}{} &
  \multicolumn{1}{l|}{CASSPR \cite{xia2023casspr}} &
  0.300 &
  0.524 &
  0.160 &
  0.427 &
  0.264 &
  0.479 &
  0.154 &
  0.416 &
  0.505 &
  0.733 &
  0.181 &
  0.411 &
  0.475 &
  0.718 &
  0.205 &
  0.469 \\
\multicolumn{1}{l|}{} &
  \multicolumn{1}{l|}{CROSSLOC3D \cite{guan2023crossloc3d}} &
  0.711 &
  0.836 &
  0.634 &
  0.846 &
  0.635 &
  0.800 &
  0.581 &
  {\color[HTML]{0000FF} 0.833} &
  {0.790} &
  {0.907} &
  0.665 &
  0.865 &
  0.738 &
  0.876 &
  0.665 &
  0.848 \\
\multicolumn{1}{l|}{} &
  \multicolumn{1}{l|}{MinkLoc3Dv2 \cite{komorowski2022improving}} &
  {0.750} &
  {0.846} &
  {0.722} &
  {0.882} &
  {0.664} &
  {0.806} &
  {0.620} &
  {0.831} & 
  {\color[HTML]{0000FF} 0.853} &
  0.931 &
  {0.742} &
  {0.896} &
  {0.801} &
  {0.895} &
  {0.737} &
  {0.885} \\
\multicolumn{1}{l|}{} &
  \multicolumn{1}{l|}{HeLiOS-S} &
  {\color[HTML]{0000FF} 0.767} &
  {\color[HTML]{0000FF} 0.867} &
  {\color[HTML]{0000FF} 0.765} &
  {\color[HTML]{0000FF} 0.912} &
  {\color[HTML]{0000FF} 0.682} &
  {\color[HTML]{0000FF} 0.815} &
  {\color[HTML]{0000FF} 0.646} &
  0.824 & 
  {0.850} &
  {\color[HTML]{0000FF} 0.934} &
  {\color[HTML]{0000FF} 0.785} &
  {\color[HTML]{0000FF} 0.919} &
  {\color[HTML]{0000FF} 0.808} &
  {\color[HTML]{0000FF} 0.905} &
  {\color[HTML]{0000FF} 0.740} &
  {\color[HTML]{0000FF} 0.892} \\  
 
\multicolumn{1}{c|}{\multirow{-8}{*}{\rotatebox{90}{Aeva}}} &
  \multicolumn{1}{l|}{HeLiOS} &
  {\color[HTML]{FF0000} \textbf{0.806}} &
  {\color[HTML]{FF0000} \textbf{0.885}} &
  {\color[HTML]{FF0000} \textbf{0.849}} &
  {\color[HTML]{FF0000} \textbf{0.940}} &
  {\color[HTML]{FF0000} \textbf{0.744}} &
  {\color[HTML]{FF0000} \textbf{0.850}} &
  {\color[HTML]{FF0000} \textbf{0.737}} &
  {\color[HTML]{FF0000} \textbf{0.883}} &
  {\color[HTML]{FF0000} \textbf{0.886}} &
  {\color[HTML]{FF0000} \textbf{0.950}} &
  {\color[HTML]{FF0000} \textbf{0.818}} &
  {\color[HTML]{FF0000} \textbf{0.930}} &
  {\color[HTML]{FF0000} \textbf{0.857}} &
  {\color[HTML]{FF0000} \textbf{0.936}} &
  {\color[HTML]{FF0000} \textbf{0.773}} &
  {\color[HTML]{FF0000} \textbf{0.903}} \\ \bottomrule
\end{tabular}%
}
\vspace{-5mm}
\end{table*}

%% file: tab/ablation_single.tex
\begin{table}[]
\centering
\caption{Performance Comparison with Homogeneous LiDAR}
\label{tab:single}
\resizebox{0.98\columnwidth}{!}{%
\begin{tabular}{lcccccc} \toprule
 &
  \multicolumn{2}{c}{\texttt{NCLT}} &
  \multicolumn{2}{c}{\texttt{R01-03} (Aeva)} &
  \multicolumn{2}{c}{\texttt{R01-03} (Livox)} \\ \cmidrule{2-7}
\multirow{-2}{*}{Method} & 
  AR@1 &
  AR@5 &
  AR@1 &
  AR@5 &
  AR@1 &
  AR@5 \\ \midrule
SOLID &
  0.217 &
  0.409 &
  0.423 &
  0.537 &
  0.331 &
  0.479 \\
PointNetVLAD &
  0.893 &
  0.966 &
  0.661 &
  0.700 &
  0.768 &
  0.830 \\
LoGG3D-Net &
  0.403 &
  0.670 &
  0.368 &
  0.368 &
  0.536 &
  0.730 \\
CASSPR &
  {\color[HTML]{FF0000} \textbf{0.961}} &
  0.993 &
  0.696 &
  0.775 &
  0.809 &
  0.877 \\
CROSSLOC3D &
  0.953 &
  {\color[HTML]{FF0000} \textbf{0.996}} &
  {\color[HTML]{0000FF} 0.759} &
  {\color[HTML]{0000FF} 0.837} &
  {\color[HTML]{0000FF} 0.835} &
  {\color[HTML]{0000FF} 0.908} \\
MinkLoc3Dv2 &
  0.941 &
  {\color[HTML]{0000FF} 0.996} &
  0.755 &
  0.820 &
  {\color[HTML]{FF0000} \textbf{0.853}} &
  {\color[HTML]{FF0000} \textbf{0.922}} \\
HeLiOS-S &
  {\color[HTML]{0000FF} 0.953} &
  0.992 &
  {\color[HTML]{FF0000} \textbf{0.777}} &
  {\color[HTML]{FF0000} \textbf{0.842}} &
  0.831 &
  0.900 \\ \bottomrule
\end{tabular}%
}
\vspace{-2mm}
\end{table}

%% file: tab/ablation_loss.tex
\begin{table}[]
\caption{Ablation Study with Loss for Each Class}
\label{tab:loss}
\centering
\resizebox{0.98\columnwidth}{!}{%
\begin{tabular}{cccrrrr} \toprule
\multicolumn{3}{c}{Loss for class} &
  \multicolumn{2}{c}{Aeva (DB)} &
  \multicolumn{2}{c}{Ouster (DB)} \\ \midrule
$(p, n)$ &
  $(s, n)$ &
  $(p, s)$ &
  \multicolumn{1}{c}{AR@1} &
  \multicolumn{1}{c}{AR@5} &
  \multicolumn{1}{c}{AR@1} &
  \multicolumn{1}{c}{AR@5} \\ \midrule
 &
  $\mathcal{L_{TSAP}}$ &
  \multicolumn{1}{c|}{$\mathcal{L_{TSAP}}$} &
  0.701 &
  0.847 &
  0.722 &
  0.854 \\
 &
  - &
  \multicolumn{1}{c|}{-} &
  0.722 &
  0.857 &
  0.756 &
  0.877 \\
 &
  $\mathcal{L_T}$ &
  \multicolumn{1}{c|}{-} &
  0.771 &
  0.885 &
  0.793 &
  0.903 \\
 &
  - &
  \multicolumn{1}{c|}{$\mathcal{L_T}$} &
  {\color[HTML]{0000FF} {0.778}} &
  {\color[HTML]{0000FF} {0.887}} &
  0.795 &
  {\color[HTML]{0000FF} {0.905}} \\
 &
  $\mathcal{L_T}$ &
  \multicolumn{1}{c|}{$\mathcal{L_T}$} &
  0.778 &
  0.885 &
  {\color[HTML]{0000FF} {0.800}} &
  0.903 \\
\multirow{-6}{*}{$\mathcal{L_{TSAP}}$} &
  $\mathcal{L_{GT}}$ &
  \multicolumn{1}{c|}{$\mathcal{L_{GT}}$} &
  {\color[HTML]{FF0000} \textbf{0.784}} &
  {\color[HTML]{FF0000} \textbf{0.890}} &
  {\color[HTML]{FF0000} \textbf{0.809}} &
  {\color[HTML]{FF0000} \textbf{0.910}} \\ \bottomrule
\end{tabular}%
} \vspace{-6mm}
\end{table}

%% file: tab/ablation_network.tex
\begin{table}[!t]
\centering
\caption{Ablation Study for Sphere Transformer Variations}
\label{tab:sphere}
\resizebox{0.98\columnwidth}{!}{%
\begin{tabular}{lcccc}
\toprule
                                       & \multicolumn{2}{c}{Aeva (DB)} & \multicolumn{2}{c}{Ouster (DB)} \\ \cmidrule{2-5} 
\multirow{-2}{*}{Method}                  & AR@1         & AR@5        & AR@1          & AR@5         \\ \cmidrule{1-5} 
w/o S.T.                                 & 0.772       & 0.883      & 0.797        & 0.904       \\
S.T. w/ $r_{f}$                                & 0.778       & 0.888      & 0.795        & 0.901       \\
S.T. w/ $r_{w}$                                 & 
{\color[HTML]{0000FF} {0.778}}    & 
{\color[HTML]{FF0000} \textbf{0.891}}    &
{\color[HTML]{0000FF} {0.801}}    & 
{\color[HTML]{0000FF} {0.906}}       \\
S.T. w/ $r_w$ (Expanding)                    &
  {\color[HTML]{FF0000} \textbf{0.784}} &
  {\color[HTML]{0000FF} {0.890}} &
  {\color[HTML]{FF0000} \textbf{0.809}} &
  {\color[HTML]{FF0000} \textbf{0.910}}\\      \bottomrule
\end{tabular} }\vspace{-1mm}
\end{table}

%% file: tab/ablation_dim.tex
\begin{table}[]
\centering
\caption{Recall and Runtime with Various Dimension}
\label{tab:dim}
\resizebox{0.98\columnwidth}{!}{%
\begin{tabular}{lccccc} \toprule
\multicolumn{1}{l}{Dimension} & \multicolumn{2}{c}{Aeva (DB)} & \multicolumn{2}{c}{Ouster (DB)} & \multicolumn{1}{c}{\multirow{2}{*}{Runtime (ms)}} \\ \cmidrule{2-5}
\multicolumn{1}{l}{$(m, l, e)$}     & AR@1   & AR@5   & AR@1   & AR@5                        & \multicolumn{1}{c}{} \\ \midrule
\multicolumn{1}{l|}{(8, 32, 0)}     & 0.715 & 0.854 & 0.758 & \multicolumn{1}{c|}{0.877} &    26.4 + 0.7        \\
\multicolumn{1}{l|}{(16, 32, 0)}    & 0.751 & 0.874 & 0.773 & \multicolumn{1}{c|}{0.890} &    26.4 + 1.2                  \\
\multicolumn{1}{l|}{(32, 64, 0)}       & 0.758 & 0.878 & 0.781 & \multicolumn{1}{c|}{0.894} &    26.4 + 5.0                  \\
\multicolumn{1}{l|}{(32, 64, 256)}  & 0.772 & \color[HTML]{0000FF} 0.886 & 0.792 & \multicolumn{1}{c|}{0.901} &   26.5 + 5.5     \\
\multicolumn{1}{l|}{(64, 128, 0)}   & \color[HTML]{0000FF} 0.778 & 0.886 & \color[HTML]{0000FF} 0.796 & \multicolumn{1}{c|}{\color[HTML]{0000FF} 0.902} &   26.5 + 19.6                  \\
\multicolumn{1}{l|}{(64, 128, 256)} &  {\color[HTML]{FF0000} \textbf{0.784}} & \color[HTML]{FF0000} \textbf{0.890} & \color[HTML]{FF0000} \textbf{0.809} & \multicolumn{1}{c|}{\color[HTML]{FF0000} \textbf{0.910}} & 26.5 + 19.8 \\ \bottomrule                   
\end{tabular}%
}
\vspace{-6mm}
\end{table}

%% file: 5_conclusion.tex
 \section{Conclusion}
\label{sec:conclusion}
In this paper, we present HeLiOS, the first deep network for heterogeneous \ac{LPR}. 
HeLiOS utilizes a local spherical transformer to learn the local distribution from each LiDAR and optimal transport-based clustering to aggregate the local features. 
Our overlap-based data mining and guided-triplet loss address the limitations of distance-based mining and fixed-margin triplet loss, resulting in more effective embeddings. 
Evaluations with public datasets show HeLiOS outperforms existing methods and demonstrates robustness in long-term place recognition with unseen LiDARs. 
Ablation studies validate the impact of loss functions, model architecture, and descriptor dimensions. 
As the first heterogeneous \ac{LPR} framework, HeLiOS opens up new opportunities for future work, including reranking tasks to enhance performance or integration with LiDAR \ac{SLAM} \cite{lee2024lidar, jung2023asynchronous, chen2024multi} and multi-robot applications \cite{huang2021disco, chang2022lamp}.
\newpage